# RUSSIAN WORD SENSE INDUCTION BY CLUSTERING AVERAGED WORD EMBEDDINGS

**Kutuzov A.B.** (andreku@ifi.uio.no)
University of Oslo, Oslo, Norway

*The paper reports our participation in the shared task on word sense induction and disambiguation for the Russian language (RUSSE'2018). Our team was ranked $2^{nd}$ for the wiki-wiki dataset (containing mostly homonyms) and $5^{th}$ for the bts-rnc and active-dict datasets (containing mostly polysemous words) among all 19 participants.*

*The method we employed was extremely naive. It implied representing contexts of ambiguous words as averaged word embedding vectors, using off-the-shelf pre-trained distributional models. Then, these vector representations were clustered with mainstream clustering techniques, thus producing the groups corresponding to the ambiguous word' senses. As a side result, we show that word embedding models trained on small but balanced corpora can be superior to those trained on large but noisy data — not only in intrinsic evaluation, but also in downstream tasks like word sense induction.*

## 1. Introducing word sense induction task

Human language is inherently ambiguous on all of its tiers. Grammatical and syntactic ambiguity is successfully solved by part-of-speech taggers and dependency parsers. But this is not enough, as morphologically and syntactically identical words can possess different senses or meanings. Indeed, all that happens with semantics, happens at the level of word senses, not words. This means that some ways of disambiguating ambiguous words and finding out the correct number of senses have to be devised.

**Word sense induction** (WSI) is an important part of computational lexical semantics and boils down to the task of automatically discovering the senses of semantically ambiguous words from unannotated text. It has long research history for English and other languages, with several relevant *SemEval* shared tasks [14]. However, until recently, the NLP community lacked proper evaluation of WSI methods for Russian. RUSSE'2018 shared task [16][1] fills in this gap. This paper describes the approach we used in the framework of this competition.

The participants of the shared task were given three sets of Russian utterances containing semantically ambiguous words. The participating systems had to group the utterances containing a particular ambiguous word into clusters, depending on the sense this word takes in this particular utterance.

The organizers offered two tracks:

1. *Knowledge-rich*, where the participants were permitted to use dictionaries or other lexical databases containing sense inventories;

2. *Knowledge-free*, where participants were allowed to use only text corpora and models automatically derived from these corpora.

We participated in the *knowledge-free* track. Thus, we had to infer word senses from the data, without relying on any external sources like *WordNet* [13], *BabelNet* [15] or *Wiktionary*[2]. The performance of the systems was evaluated by calculating the Adjusted Rand Index (ARI) between context clustering produced by the systems and the gold clustering provided by the organizers.

We intentionally employed a very simplistic (even naive) approach to word sense induction, which we describe below. The reason for this was that we were interested in whether Russian WSI task can be solved using only already available algorithms and off-the-shelf models. It turned out to be true for one of the three RUSSE'18 datasets (we ranked 2[nd]) but not so true for other two (we ranked 5[th]). It should be noted, however, that none of the participants achieved reasonably high scores for these last 2 datasets. We outline the differences between the datasets in Section 3.

Overall, our contributions are twofold:

1. We describe and publish the WSI system for Russian, which produces very competitive results for homonyms with non-related senses, and which is based exclusively on off-the-shelf tools and models.

1 https://russe.nlpub.org/2018/wsi/
2 https://ru.wiktionary.org

2. It was already known that training corpus balance can be even more important for word embedding models than its size, when evaluated intrinsically. In this paper, we show that this holds for extrinsic evaluation setting as well, with WSI as a downstream task in this case.

The rest of the paper is organized as follows. Section 2 briefly outlines the previous work related to word sense induction and distributional semantics. In Section 3, we present the datasets offered by the shared task organizers and the corpora used to train our word embedding models. Section 4 provides the details of the employed approach. In Section 5 we describe the results, comparing them to other participants, and in Section 6 we conclude.

## 2. Related Work

Word sense induction task is closely related to word sense disambiguation: the task to assign meanings to ambiguous words from a pre-defined sense inventory. Even this easier task is notoriously difficult to handle computationally. In 1964, Yehoshua Bar-Hillel, Israeli mathematician and linguist even proclaimed that "'sense ambiguity could not be resolved by electronic computer either current or imaginable" [1].

Fortunately, it turned out that things are not that bad. Since the sixties, many word sense disambiguation techniques appeared, which were quite successful in telling which sense the particular word is used in. In the recent years, the majority of these techniques are based on statistical approaches and machine learning.

However, all word sense disambiguation approaches suffer from the same problem known as knowledge acquisition bottleneck. They need ready-made sense inventory for each ambiguous word: otherwise, there is nothing to choose from. Manually annotated semantic concordances and lexical databases quickly get outdated. They don't keep up with the changes in language, and humans simply cannot annotate that fast. This is especially true for named entities and for specialized domains.

At the same time, it is relatively easy to compile large up-to-date corpora of unannotated text. It is then possible to infer word sense inventories from these corpora automatically. This task is called *unsupervised word sense disambiguation* or *word sense induction* (WSI): the input is corpus, and the output consists of sense sets for each content word in the corpus we are interested in. Quoting Adam Kilgariff in [6], "*word senses are abstractions from clusters of corpus citations*".

Thus, there are no pre-defined sense inventories: we discover senses for a given word directly from text data. This boils down to the task of clustering occurrences of the input word in the corpus, based on their senses.

The foundations for clustering-based WSI were laid in [5] and [19]. In its essence, it is a very straightforward approach based on word distributions:
1. Represent each ambiguous word with a list of its context vectors;
    - context vector contains identifiers of context words in a particular context (sentence, phrase, document, etc...).
2. For each word, cluster its lists into a (predefined) number of groups, using any preferred clustering method;
3. For each cluster, find its centroid;
4. These centroids serve as sense vectors for the subsequent word sense disambiguation.

At test time, the system is given a new context (for example, sentence) containing an ambiguous input word. It computes its context vector by listing the context words, and then chooses the sense vector most similar to the current context vector.

Of course, by the nature of the approach, the induced "senses" are coarse, nameless and often not directly interpretable (see [17] for an attempt to overcome non-interpretability). However, it is still possible to tell one sense from another in context, and this is what real-world systems need. Further on, the WSI approaches were enriched with additional techniques, for example with lexical substitution [22]. Today, WSI is extensively relied upon in many NLP tasks, including machine translation and information retrieval [14].

We use prediction-based word embedding models of lexical semantics as the source of distributional information representing word meanings. This sort of models is extensively described elsewhere. See [12] and [2] for the background of *Continuous Skipgram* and *fastText* algorithms that we employed.

Note that there are many other WSI algorithms, including graph-based approaches. We refer the interested reader to [3] for the general overview and to [10] for an example of the application of graph-based WSI for Russian data. Very recent experiments with combining graph and word embedding approaches to WSI are described in [23].

## 3. Data overview

In this section, we describe the RUSSE'18 datasets, and the word embedding models we used to process them.

RUSSE'18 shared task offered three datasets (with a training and a test part in each):

1. **wiki-wiki:** sense inventories and contexts from the Russian Wikipedia articles

2. **bts-rnc**: sense inventories from "*Bolshoi Tolkovii Slovar*" dictionary (BTS), contexts from the Russian National Corpus [15]

3. **active-dict**: sense inventories from the *Active Dictionary of the Russian Language*, contexts from the examples in the same dictionary.

Each training set consisted of several ambiguous query words (from 4 in the **wiki-wiki** to 85 in the **active-dict**) and about a hundred contexts for each of them. The context as a rule included several sentences, not more than 500-600 characters total. Each context was annotated with the identifier of the sense in which the corresponding query word was used in this context. The test sets featured the same structure, of course without the sense annotation. Thus, the task was to find out for each query word in the test set how many senses it has and which contexts belong to the same senses.

The systems' performance for each dataset was evaluated separately. We strongly support this decision of the organizers and argue that it might even make sense to cast this as two independent shared tasks.

The reason is that **wiki-wiki** dataset is substantially different from the other two. First, its sense structure is much more stable: the training set query words have exactly two senses each. At the same time, for the **bts-rnc** training set the average number of senses per query word is 3.2, and the maximum number of senses is as high as 8. The **active-dict** training set is even more varied, with the average number of senses 3.7, and the maximum number of senses 17 (*sic!*).

As if this was not enough, the nature of these senses is unsurprisingly different. In the **wiki-wiki** dataset, most senses are homonyms, that is unrelated to each other (for example, "бор" *pine wood* and "бор" *boron*). On the contrary, the other two datasets are abundant in polysemy, where word senses are somehow related. Cf. "обед" *lunch* and "обед" *lunchtime* from the **bts-rnc** dataset, or "дерево" *tree* and "дерево" *wood* from the **active-dict** dataset. There are also many cases of metonymy and other subtle semantic shifts.

Of course, word senses are a kind of continuum, and there is no distinct boundary between homonymy and polysemy. Even for human experts, it is often difficult to tell how many senses does a word really have. However, we still think that the **wiki-wiki** dataset presents a very different task. This task (*inducing meanings of homonyms*) is much easier than the task of *inducing different senses of polysemous words*. Arguably, considerably different approaches are needed for both.

Anyway, to handle semantic phenomena, one needs a way to model semantic similarities and dissimilarities between words. To this end, we employed pre-trained word embedding models for Russian, downloaded from the *RusVectōrēs*[3] web service [8]. We tested five models:

1. **ruscorpora_upos_skipgram_300_5_2018** trained on the Russian National Corpus (RNC) [18] (about 250 million words);

2. **ruwikiruscorpora_upos_skipgram_300_2_2018** trained on concatenation of the RNC and the Russian Wikipedia (about 600 million words);

3. **news_upos_cbow_600_2_2018** trained on a large Russian news corpus (about 5 billion words);

4. **araneum_upos_skipgram_300_2_2018** trained on the *Araneum Russicum Maximum* web corpus [26] (about 10 billion words);

---

3 http://rusvectores.org/ru/models/

5. **araneum_none_fasttextskipgram_300_5_2018** trained on the same corpus as the previous model, but using the *fastText* algorithm instead of the *Continuous Skipgram*.

With these components at hand, we aimed to build a system capable of inducing word senses for the three datasets. In the next section, we describe this system.

## 4. Our approach

We applied more or less the same workflow for all the three datasets, with minor alterations, depending on what worked best. Briefly, our approach can be summarized in the following steps:

1. Lemmatize and PoS-tag contexts;
2. Represent each context as a fixed-length vector manifesting its semantics;
3. Determine the number of clusters in the set of contexts, using the *Affinity Propagation* algorithm;
4. Group the contexts into clusters representing word senses, using either the same *Affinity Propagation* or other clustering algorithm.

There are two important and practically independent phases in this workflow, which we describe in the next 2 subsections.

### 4.1 Contexts representations

The first phase consists of converting context utterances from lists of words to fixed length vector representations. Note that first we lemmatized and PoS-tagged all words in the context utterances using *UDPipe* 1.2 tagger [21] trained on Russian Universal Dependencies corpus [4]. We also tried to use *Mystem* tagger [20] instead, but this did not result in any improvements for the WSI task. The ambiguous query words themselves were removed from the utterances.

Then, for each lemmatized context utterance, we created "semantic fingerprints" as described in [7]. The "fingerprint" function takes as an input the list of lemmas and a pre-trained word embedding model. It looks up the embeddings for all the lemmas from the context utterance present in the model's vocabulary. Then, these vectors are averaged to produce the function output, which is a single vector of the same dimensionality as the vectors in the employed model (we used the models with the vector size 300). This dense vector is used as a semantic representation of the context utterance.

Note that we slightly modified the "semantic fingerprint" notion from [7]. First, we counted multiple occurrences of the same lemma as one occurrence (that is, binary bag-of-words was used, discarding local word frequencies in the context utterances). Second, before averaging the word vectors, we assigned them weights in the range of [0...1], in inverted proportion to the word frequencies in the training corpus of the underlying word embedding model. This way, "globally frequent" words (which are often not sense-specific) got less influence on the resulting semantic fingerprints, while "globally rare" words (often specific for a particular sense) became more influential. In our experience, both changes improved the word sense induction performance (see Section 5).

With the vector representations of contexts ("semantic fingerprints") ready, it is possible to cluster them into groups corresponding to senses of the query word.

### 4.2 Contexts clustering

Theoretically, any clustering algorithm can be used in this case. The only complication is that the number of senses (and thus the number of clusters) for any given query word is unknown. This number must be induced from the data.

Many clustering techniques are able to do this. We employed the *Affinity Propagation* algorithm: first, because it is readily accessible in the *scikit-learn* library[4], and second, because it was successfully applied to related tasks (in [22] for English and in [9] for Russian).

*Affinity Propagation* produces clustering of the contexts, which can be used immediately as the desired sense-specific grouping. For the **wiki-wiki** dataset, this was our strategy. However, for two other datasets, we found that our system performs better if we use *Affinity Propagation* only to induce the number of clusters (senses). After that, another clustering algorithm (either *K-Means* or spectral clustering) is called to separate the data into the induced number of groups. This once again emphasizes the differences between the datasets in the shared task.

---

4 We also tested DBSCAN clustering algorithm, but it yielded suboptimal results for all datasets.

Note that the *Affinity Propagation* takes two parameters: preference and damping, which both greatly influence the behavior of the algorithm, especially the resulting number of clusters. We performed grid search to determine the best combination of these parameters for each dataset[5].

The resulting system, despite its simplicity, produces reasonable clusterings. We illustrate this with the Figure 1 which presents the 2-dimensional *t-SNE* projection of 300-dimensional context vector representations for the query word "*бор*" mentioned above. Stars stand for the contexts annotated with the "*pine wood*" sense, and circles for the "*Boron*" sense. Colors reflect the clustering produced by the system. One can see that it successfully detected the correct number of clusters (2) and correctly grouped all the contexts, except one.

## 5. Results

We first present the results of our experiments on the training data, and then describe the performance of the presented system on the test sets in comparison with other participants of the shared task.

As mentioned before, we experimented with five pre-trained word embedding models. The Table 1 provides an overview of the best results that we got for each dataset using each particular model as the source of knowledge about word meanings.

It is clear that the model trained exclusively on the Russian National Corpus (RNC) was the best for all three datasets, despite comparatively small size of the corpus. This further supports the importance of proper compiling and balancing the training corpora for word embedding models. It was previously shown in [11] that the models trained on the RNC are very often not worse or even better than those trained on a much larger web corpus in the intrinsic evaluation (semantic similarity task). The present work continues this line of research and proves that this holds at least for some extrinsic evaluation settings as well (WSI in this case).

---

[5] The best values for the preference parameter seem to lie between -0.6 and -0.7, while for the damping parameter the sweet spot is between 0.7 and 0.8.

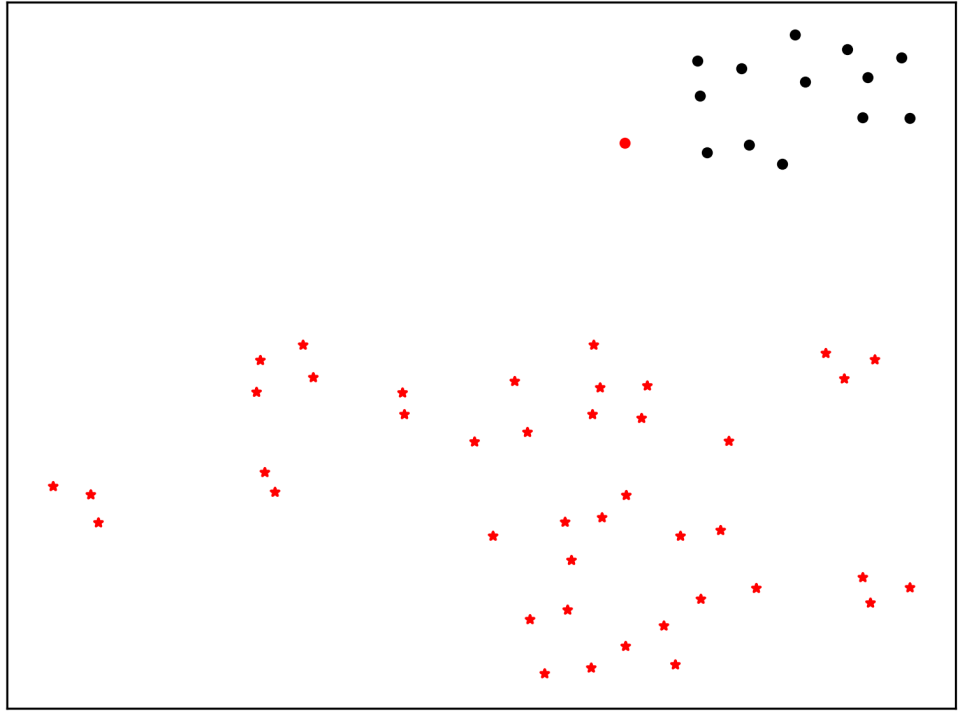

*Figure 1: Clustering of the "бор" contexts ("pine wood" and "Boron"). Colors are clusters assigned by the system, shapes are gold clusters.*

The way word vectors are averaged to produce "semantic fingerprints" greatly influences the results for the **wiki-wiki** dataset, as shown in Table 2. Changing the representation to binary bag-of-words instead of count bag-of-words brings stable improvements, as well as introducing global frequency weights. The other 2 datasets are almost agnostic to these parameters: as we believe, precisely because of their different nature. Note also that due to the usage of the second clustering algorithm (dependent on random initialization), the results for the **bts-rnc** and **active-dict** datasets are non-deterministic and fluctuate slightly from one run to another.

| Model / Dataset | wiki-wiki | bts-rnc | active-dict |
|---|---|---|---|
| *ruscorpora_upos_skipgram_300_5_2018* | **0.772** | **0.176** | **0.260** |
| *ruwikiruscorpora_upos_skipgram_300_2_2018* | 0.669 | 0.162 | 0.210 |
| *news_upos_cbow_600_2_2018* | 0.653 | 0.174 | 0.143 |
| *araneum_upos_skipgram_300_2_2018* | 0.492 | 0.162 | 0.197 |
| *araneum_none_fasttextskipgram_300_5_2018* | 0.695 | 0.171 | 0.178 |

*Table1: Clustering performance (ARI) on the training sets, depending on the pre-trained word embedding model*

| Dataset | Original semantic fingerprints | + binary bag-of-words (discarding local word frequencies) | + weights (global word frequencies) |
|---|---|---|---|
| **wiki-wiki** | 0.579 | 0.717 | **0.772** |
| **bts-rnc** | 0.169 | 0.167 | 0.176 |
| **active-dict** | 0.250 | 0.254 | 0.260 |

*Table2: Clustering performance (ARI) depending on the parameters of word vector averaging*

Finally, the Table 3 presents our scores on the test sets, and thus, the resulting performance of the presented system. To cut it short, our naive approach turned out to be very competitive for the WSI on homonyms from the **wiki-wiki** dataset, winning the 2[nd] place in the ranking with the ARI of 0.71.

For more subtle inter-related senses of the **bts-rnc** and **active-dict** datasets, our approach performed much worse, although still allowing us to stay in the top 25% results. Note that for these two datasets, none of the competing systems managed to achieve ARI higher than 0.34, which is a long way to any possible production usage. Partly this may be caused by flaws in the gold data itself: it would be an interesting research to measure human performance and inter-rater reliability in clustering contexts for these two datasets. It is quite probable that it will turn out to be not much higher.

It is also interesting that the best results for the **wiki-wiki** (including ours) and **bts-rnc** datasets outperform state-of-the-art WSI results for English, which achieve ARI about 0.215-0.286 [24, 25]. Certainly, this can be caused by the differences between the RUSSE'18 datasets and those of SemEval-2013 and WWSI, but still this phenomenon deserves a deeper analysis in the future.

|  | Our ARI ("RusVectores" team) | Rank (of 19 participants) | The best participant ARI |
| --- | --- | --- | --- |
| **wiki-wiki** | 0.7096 | 2 | 0.9625 |
| **bts-rnc** | 0.2415 | 5 | 0.3384 |
| **active-dict** | 0.2144 | 5 | 0.2477 |

*Table3: Overall shared task results (evaluated on the test sets)*

## 6. Conclusions

This is the description of our participation in the RUSSE'18 Russian Word Sense Induction shared task. We intended to create a very naive WSI system making use of pre-trained word embedding models and standard clustering algorithms. This enterprise was successful for the **wiki-wiki** dataset, but not so much for the **bts-rnc** and **active-dict** datasets: most probably, because they mostly consist of polysemous words with highly inter-related senses.

We showed that word embedding models trained on well-balanced and clean corpora (like the Russian National Corpus) can be superior in the extrinsic WSI task to those trained on large but noisy and unbalanced web or news corpora. This goes in line with the previous research which proved this for various intrinsic evaluation tasks.

The system we implemented is described in detail in this paper, and its Python source code is available online[6]. We hope that it will be of some use to other Russian NLP practitioners. Finally, we express our gratitude to the RUSSE'18 organizers for the chance to participate in an exciting shared task.

---

6 https://github.com/akutuzov/russian_wsi